\pdfoutput=1 
\def\year{2020}\relax
\documentclass[letterpaper]{article} 
\usepackage{aaai20}  
\usepackage{times}  
\usepackage{helvet} 
\usepackage{courier}  
\usepackage[hyphens]{url}  
\usepackage{graphicx} 
\usepackage{amsmath}
\usepackage{graphicx}  
\usepackage{multirow}
\usepackage{hyperref}

\makeatletter
\newcommand{\printfnsymbol}[1]{%
  \textsuperscript{\@fnsymbol{#1}}%
}
\makeatother

\newcommand{\citet}[1]{\citeauthor{#1} \shortcite{#1}}
\newcommand{\citep}{\cite}

\title{AWR: Adaptive Weighting Regression for 3D Hand Pose Estimation}
\author{
    Weiting Huang,\textsuperscript{\rm 1,2}\thanks{The authors have equal contribution to the work and are listed in alphabetical order.} 
    Pengfei Ren,\textsuperscript{\rm 1,2}\printfnsymbol{1} 
    Jingyu Wang,\textsuperscript{\rm 1,2}\thanks{Corresponding authors.} 
    Qi Qi,\textsuperscript{\rm 1,2}\printfnsymbol{2} 
    Haifeng Sun\textsuperscript{\rm 1,2}\printfnsymbol{2} \\
    \textsuperscript{\rm 1}State Key Laboratory of Networking and Switching Technology, \\ Beijing University of  Posts and Telecommunications, Beijing 100876, P.R. China \\
    \textsuperscript{\rm 2}EBUPT Information Technology Co., Ltd., Beijing 100191, P.R. China \\
    \{huangweiting, renpengfei, wangjingyu, qiqi, sunhaifeng\}@ebupt.com
}
\begin{document}

\maketitle

\begin{abstract}
In this paper, we propose an adaptive weighting regression (AWR) method to leverage the advantages of both detection-based and regression-based method. Hand joint coordinates are estimated as discrete integration of all pixels in dense representation, guided by adaptive weight maps. This learnable aggregation process introduces both dense and joint supervision that allows end-to-end training and brings adaptability to weight maps, making network more accurate and robust. Comprehensive exploration experiments are conducted to validate the effectiveness and generality of AWR under various experimental settings, especially its usefulness for different types of dense representation and input modality. Our method outperforms other state-of-the-art methods on four publicly available datasets, including NYU, ICVL, MSRA and HANDS 2017 dataset. 

\end{abstract}

\section{Introduction}

Accurate and robust 3D hand pose estimation is crucial in applications of human-computer interaction, such as virtual reality and augmented reality. The accuracy and speed have improved significantly in recent years thanks to the emergence of deep cameras and deep learning. However, it is still challenging to achieve accurate and robust results due to the low quality of depth images, extreme viewpoints,  severe self-occlusion and self-similarity among fingers.

\parskip=0pt

Recently, deep neural network based methods have shown great performance in 3D hand pose estimation \cite{nyu,p2p,srn,ren,pose,v2v,dense3d,a2j,pixelwise,3dcnn}. These works can be divided into two categories according to networks' output type — regression-based methods and detection-based methods. Regression-based methods directly map input depth images to 3D hand pose parameters or hand joint coordinates. They allow different regions on feature maps to be weighted in the fully connected layers to regress joint coordinates. Thus it captures global constraints among hand joints and performs better in extreme viewpoints where severe occlusion occurs \cite{vanora}. However, fully connected layers flatten the feature map and destroy the spatial structure in it, making it hard to learn the highly non-linear mapping from image space to joint space. 

\parskip=0pt

\begin{figure}[t]
\centering
\includegraphics[width=1\columnwidth]{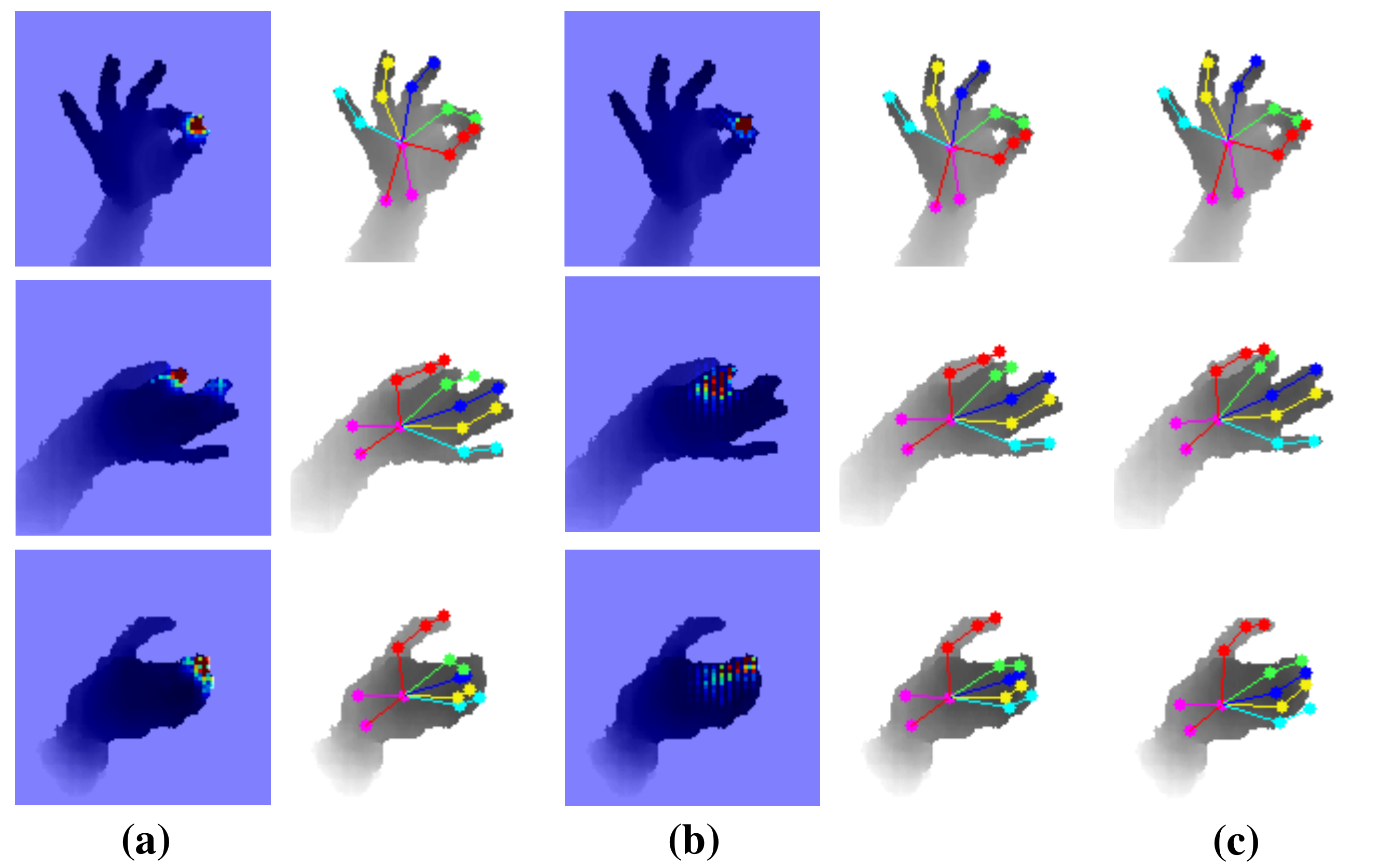} 
\caption{Weight maps produced during aggregation for (a) standard detection-based methods and (b) our method, followed by their estimation results and (c) corresponding ground truth labels under situations where the target joint (index fingertip) is visible (top row) or occluded (middle row) and when there exists self-similarity among fingers (bottom row). Figure is best viewed in color.}
\end{figure}

\parskip=0pt

Detection-based methods instead generate dense estimations such as heatmaps\cite{point2pose,simple} or offset vector fields\cite{dense3d,srn}, then the joint coordinates are inferred through a non-learnable information aggregation process such as the taking argmax operation or mean-shift estimation, guided by weight maps. The fully convolutional network architecture keeps the spatial structure of extracted feature maps and better exploits local evidence in depth images. However, the information aggregation is treated as post processing and separated from network training, posing gaps between training and inferencing. Besides, detection-based methods force networks to learn a fixed range of weight distribution and strictly restrict the post processing information aggregation process. Specifically, ground truth heatmaps are generated by a fixed size Gaussian blobs for heatmaps based methods. And offset vector fields based methods must strictly limit the number \cite{dense3d} or the distribution range of candidate points \cite{p2p} in order to prevent points which are irrelevant to target joint from affecting estimation accuracy. The above procedure reduces the generalization ability of networks and makes estimations unstable when depth values near the target joint are heavily missing.

\parskip=0pt

In this paper, we propose a simple yet effective adaptive weighting regression (AWR) method to unify the two complementary lines of work in a single pass. Guided by adaptive weight maps, AWR aggregates different regions of dense representation through discrete integration of all pixels in it. This operation is differentiable so that it can be embedded into the network for end-to-end training and applies direct supervision on joint coordinates, drawing consensus in network's supervision and output. The inferred joint coordinates are continuous and up to arbitrary accuracy. Besides, the weight distribution in weight maps can be adjusted adaptively to achieve more accurate and robust performance under the guidance of joint supervision. As shown in Fig 1, when target joint is visible and easy to distinguish, the weight distribution of AWR tends to focus more on pixels around it as standard detection-based methods do, which helps to make full use of local evidence. When depth values around the target joint are heavily missing due to occlusion or under the situation of severe self-similarity among fingers, the weight distribution spreads out to capture information of adjacent joints. This mechanism makes the network more robust to situations where depth values around hand joints are heavily missing. Therefore, AWR greatly improves network's accuracy and robustness through learnable aggregation and shares the merits of both regression-based and detection-based methods. 

\parskip=0pt

The idea of aggregating different regions in dense representations to derive joint coordinates has previously been seen in human \cite{integral} or hand pose estimation \cite{a2j,pixelwise}. However, \cite{integral} for human pose estimation focus on RGB images and this type of representation ignores the 3D geometric properties of essentially 2.5D depth images. And \cite{a2j,pixelwise} for hand pose estimation are specialized for certain type of network structure, dense representation and input modality. The effectiveness and generality of this operation are not extensively validated. We present a simple yet effective network to empirically show the improvement in accuracy and robustness brought by adaptive weighting regression. This helps to inspire and pave the way for future works in 3D hand pose estimation.

\parskip=0pt

We conduct comprehensive explorations to illustrate the effectiveness and generality of our proposed AWR under various experimental settings, including \textit{different types of dense representation, input modality, network architecture as well as input and dense representation sizes}. Experiments show that AWR brings improvements in both accuracy and robustness. The pipeline of our method and arrangements for experiments can be seen in Fig 2. Our proposed method outperforms previous state-of-the-art methods on four publicly available datasets, including NYU \cite{nyu}, ICVL \cite{icvl}, MSRA \cite{offset2} and HANDS 2017 \cite{hands17} dataset. 

\parskip=0pt

Our main contribution is presenting an adaptive weighting regression (AWR) method to aggregate dense representation through discrete integration. AWR unifies the dense representation and hand joint regression to enable direct supervision on joint coordinates, narrowing the gap between training and inferencing. Comprehensive explorations have been done to validate the improvement in network's accuracy and robustness brought by AWR as well as its generality to work under various experimental settings. The overall network is simple yet effective and achieves state-of-the-art performance on four publicly available datasets. Code is available at \url{https://github.com/Elody-07/AWR-Adaptive-Weighting-Regression}.

\parskip=0pt

\begin{figure}[t]
\centering
\includegraphics[width=1\columnwidth]{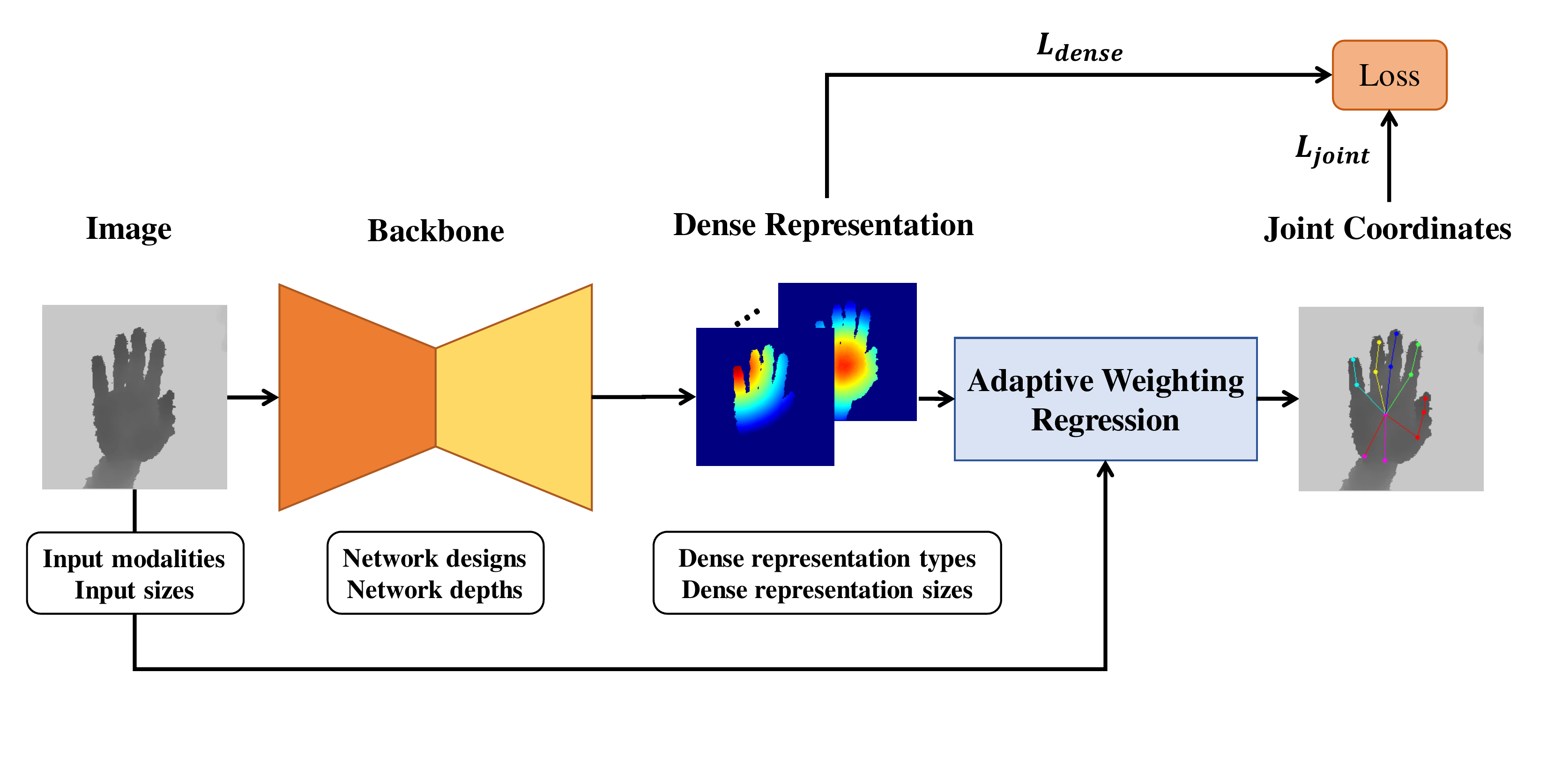} 
\caption{The pipeline of our method and arrangements for comprehensive explorations. Figure is best viewed in color.}
\end{figure}

\section{Methodology}
We formulate hand pose estimation as a dense representation embedded regression problem to leverage the advantages of both regression-based and detection-based methods. We utilize fully convolutional 2D CNN to generate dense representation, enabling network to maintain the spatial structure of extracted feature maps and better exploit local pattern in depth images. Then a learnable and adaptive weighting operation is applied to aggregate spatial information of different regions in dense representation. It has two advantages, including 1) adding direct supervision on joint coordinates to draw consensus between training and inferencing; 2) enhancing network's robustness and generalization ability by adaptively aggregating spatial information from related regions in dense representation.

\parskip=0pt

\subsection{Adaptive Weighting Regression}
Adaptive weighting regression (AWR) aggregates different regions in dense representation by taking discrete integration of all pixels on it to derive joint coordinates, guided by adaptive weight maps. Weight maps are crucial for AWR since they help network focus more on regions related to the target joint. To achieve accurate and robust estimation, they need to have the following two properties. Firstly, during aggregation, since points around the target joint give more accurate predictions of joint coordinates \cite{point2pose}, weights of these points need to be higher. The distance information contained in heatmaps or offsets can be considered as a natural depiction of this property. Secondly, when depth values around hand joints are heavily missing, weight distribution should spread out to more related regions to consider correlations among hand joints and enhance the network's robustness. Most detection-based methods adopt information aggregation during post processing. However, the post processing is not learnable and the generated weight maps cannot adapt to different situations, impairing the network's generalization ability to occlusions or severe self-similarity among fingers. 

\parskip=0pt

We present an adaptive weighting regression operation to transform dense spatial representation into joint coordinates. It is a discrete integral operation and can be described as follows:

\parskip=0pt

\begin{small}
\begin{equation}
\begin{aligned}
p_j = \sum_{i=1}^{n}w_j(p_{ij})\times j(p_{ij})
\end{aligned}
\end{equation}
\end{small}

\noindent where $p_{ij}$ denotes one of $n$ pixels in dense representation of joint $j$; $j(p_{ij})$ denotes hand joint coordinates recovered from point $p_i$ for joint $j$; $w_j(p_{ij})$ denotes the value of point $p_{ij}$ in weight maps, normalized using softmax function. The operation is differentiable and computationally simple but unifies the dense representation and hand joint coordinates.

This operation is learnable and allows end-to-end training, which narrows the gap between training and inferencing. Benefiting from both dense representation and joint coordinates supervision, our method is able to adaptively aggregate spatial information in dense representation. Besides, it is simple, fast and adding negligible overhead in computation and storage.

\parskip=0pt

\subsection{Comprehensive Explorations}
The main purpose of our work is to explore the effectiveness and generality of adaptive weighting regression (AWR) method. Therefore, we conduct extensive explorations under various experimental settings, including \textit{dense representation types, input modalities, network architectures as well as input and dense representation sizes}. 

\parskip=0pt

\textbf{Dense representation types.} We research on currently most frequently used dense representation types in pose estimation and select six of them to observe the performance of AWR. Firstly, following \cite{point2pose}, we propose a simple dense representation by stacking x, y and z coordinate of the target joint. Specifically, each joint corresponds to a pose map with three channels where each channel contains the same value of x, y or z coordinate of the joint. Since pose heatmaps lack the measurement of distance from pixels to target joint, weight maps are learned without supervision to guide the aggregation of local evidence. This type of representation is referred to as "P".

Heatmaps are the most commonly used representation in human and pose estimation, especially probability heatmaps where each pixel represents the probability of it being a specific joint. Since probability heatmaps only represent 2D properties of joint coordinates and 3D heatmaps are computationally expensive, we introduce depth maps or depth offset maps to encode depth information. Specifically, the depth maps and depth offset maps have the same resolution as probability heatmaps and denotes the depth value of target joint or depth offset from current pixel to target joint respectively. We refer to probability heatmaps combined with depth maps as "H1" and the other as "H2". 

Recent work \cite{a2j} uses 2D offsets between anchor points and hand joints to represent 2D positions of joints. Due to the large variance of offsets, we further decompose them into 2D directional unit vector fields and closeness heatmaps, reflecting 2D directions and closeness from each pixel in depth images to target joints. Similarly, 2D offsets lack the prediction in depth. Therefore, we simultaneously predict depth value for each pixel or depth offset from current pixel to target joint and they are referred to as "O1" and "O2". 

The representation of offsets is continuous and up to arbitrary localization accuracy. But separating the prediction of plane and depth coordinate and using only 2D distance to generate closeness heatmaps ignores the 3D property of depth images to some extent. 3D offsets, as used in \cite{srn,dense3d}, instead predict 3D directional unit vector fields and closeness heatmaps, reflecting 3D directions and per-pixel closeness to target hand joint. They unify the prediction of three coordinates together and fully exploit the 3D spatial information present in depth images. We refer to 3D offsets as "O3" in the experiment section.

\parskip=0pt

The formulation of offset representation $\phi(p_i, p_j)$ is shown in Eq 2. 

\begin{small}
\begin{equation}
    \phi(p_i, p_j)=\left\{
        \begin{array}{ll}
            1_{Hand}(p_i)\times(p_i-p_j), & \textrm{$||p_i-p_j|| \leq k$}\\
            0, & \textrm{otherwise}
        \end{array} 
        \right.
\end{equation}
\end{small}

\parskip=0pt

\noindent where $p_i$ and $p_j$ denote depth, 2D or 3D coordinates of a pixel in depth images and target hand joint respectively; $k$ denotes the maximum distance from pixels in depth images to target hand joint and $1_{Hand}(p_i)$ is an indicator function. If $p_i$ belongs to hand, $1_{Hand}(p_i)$ equals to 1 otherwise 0.

\parskip=0pt

For 2D and 3D offsets, they are further decomposed into 2D or 3D directional unit vectors $V(p_i, p_j)$ and closeness heatmaps $S(p_i, p_j)$.  

\begin{small}
\begin{equation}
\begin{aligned}
    S(p_i, p_j) &= \left\{
        \begin{array}{ll}
            1_{Hand}(p_i)\times\frac{k-||p_i-p_j||}{k},&\textrm{$||p_i-p_j||\leq k$} \\
            0,&\textrm{otherwise}
        \end{array}
        \right.\\
    V(p_i, p_j) &= \left\{
        \begin{array}{ll}
            1_{Hand}(p_i)\times\frac{p_i-p_j}{||p_i-p_j||},&\textrm{$||p_i-p_j||\leq k$} \\
            0, & \textrm{otherwise}
        \end{array}
        \right.
\end{aligned}
\end{equation}
\end{small}

\parskip=0pt

\textbf{Input modalities.} To fully exploit 3D geometric information in depth images, recent researchers tend to transform depth images into 3D voxels \cite{v2v} or point clouds \cite{handpointnet,p2p} and apply a 3D CNN or PointNet \cite{pointnet}. We adopt three different input modalities (depth images, voxels and point clouds) in our experiments to testify the generality of AWR. For voxels, we follow the pipeline in \cite{v2v} which transform the depth images into occupancy voxels and apply 3D hourglass network structure, except that our network outputs 3D offsets to hand joints, which are further decomposed into 3D directional unit vectors and 3D closeness heatmaps. We select voxels whose distance to the target joint is within 15 voxels as candidate voxels to recover joint coordinates from. As for point clouds, we adopt a similar network structure as in \cite{p2p} and predict 3D offsets, except that to have a fair comparison with other input modalities, we do not pre-process the point clouds, neither transform the point clouds into oriented bounding box (OBB) coordinate or refine fingertips. 

\parskip=0pt

\textbf{Network architectures.} We use simple 2D CNN network which contains only a backbone network, such as ResNet\cite{resnet} or Hourglass\cite{hourglass} to extract feature maps. Then several deconvolutional layers are applied to lift the resolution of extracted feature maps, followed by a few convolutional heads to generate dense representation. Since different types of dense representation may consist of independent components with different physical characteristics, we use separated convolutional heads to generate them. Specifically, for O3, two heads are used to output directional unit vector fields and closeness heatmaps respectively. While for O1 and O2, we add another head to generate depth maps or depth offset maps. For H1 and H2, two convolution heads are applied to output probability heatmaps and depth maps or depth offset maps respectively. For P, the two heads output pose and weight maps respectively and weight maps are adaptively learned by the network without supervision.

We research on several backbone networks to show the effectiveness and generality of AWR. Specifically, network designs of ResNet \cite{resnet} and multi-stage Hourglass \cite{hourglass} and network depths of ResNet18, 50, 101 are investigated. Results show that AWR corporates well with various network structures.

\parskip=0pt

\textbf{Input and dense representation sizes.} The performance of standard detection-based methods is affected by the resolution of inputs and dense representations. To attain higher accuracy, larger input and dense representation sizes are required. We try out different sizes for both input depth images and dense representation to show the robustness of our method to resolution change. We show that even with small input and dense representation resolution, our method still achieves considerable good performance, which makes our method favorable in real-world applications when computational cost is restricted. 

\parskip=0pt

\subsection{Implementation Details}

There are two supervisions in our method: dense representation supervision and joint coordinates supervision. We first pre-train the network using only dense representation supervision, then finetune the network with joint supervision. We find that this pipeline works better than training with both supervisions simultaneously. We argue the reason is that the former provides weight maps with more adaptability. Following previous works \cite{srn}, we adopt smooth L1 loss for joint supervision since it is less sensitive to outliers than L2 loss. However, dense representation loss varies for different dense representation types. We try out smooth L1 loss as well as L2 loss to find better fitting loss type and it turns out that for offsets, smooth L1 loss performs better, while for heatmaps, L2 loss is more stable. 

\parskip=0pt

Our method is implemented with PyTorch using Adam \cite{adam} optimizer with initial learning rate of 0.001 and weight decay of 0.0005. The batch size is set to 32. We train the network with initial learning rate and then drop as the performance reaches a plateau. Same as \cite{srn}, we first train a small separated 2D CNN to attain hand center and extract hand regions from depth images, then adopt data augmentation strategies in world coordinate, to prevent the network from overfitting, including random rotation ([-180, 180]), random translation ([-10, 10]) and random scaling ([0.9, 1.1]).

\parskip=0pt

\section{Experiments}

\subsection{Datasets and Evaluation Metrics}

We conduct experiments on four publicly available hand pose datasets: NYU dataset \cite{nyu}, ICVL dataset \cite{icvl}, MSRA dataset \cite{offset2} and HANDS 2017 dataset \cite{hands17}. \textbf{NYU dataset} contains 72K and 8K frames for training and evaluation respectively. Each frame contains 3D ground truth coordinates for 36 joints. Following previous works in \cite{p2p}, we apply our method on a subset of 14 hand joints. \textbf{ICVL dataset} consists of 330K training frames and 2 testing sequences with each 800 frames. The dataset is collected from 10 different subjects with 16 hand joint annotations for each frame. \textbf{MSRA dataset} contains 76.6K frames and there are 21 hand joint annotations in each frame. We adopt the leave-one-subject-out cross-validation strategy for MSRA dataset. \textbf{HANDS 2017} dataset is currently the largest dataset with 957K training and 295K testing frames. The dataset contains 21 hand joint annotations for each frame.

\parskip=0pt

Since NYU dataset has a relatively wider coverage of hand poses and its test set labels are publicly available, we conduct exploration experiments on NYU dataset to investigate the effectiveness and generality of our proposed AWR under various experimental settings. The other three datasets are mainly used for comparison with previous state-of-the-art methods. 

\parskip=0pt

We evaluate our method using two commonly used metric. The first metric is per-joint and all-joint mean error over all test frames. The joint error is computed as Euclidean distance between predicted and ground truth hand joint. The second metric is the proportion of good frames over all test frames. A frame is considered good only when each joint's error is within a threshold.

\begin{table}[t]
\centering
\caption{Results of six types of dense representation trained with only dense representation supervision (Dense), only joint supervision (Joint) and both supervisions (Both) on NYU dataset. R represents directly regressing joint coordinates using ResNet50.}
\resizebox{.8\columnwidth}{!}{
\begin{tabular}{c|c|c|c}
\hline
\multirow{2}{*}{\textbf{Representation}} & \multicolumn{3}{|c}{\textbf{Mean Error (mm)}} \\ \cline{2-4}
 & \textbf{Dense} & \textbf{Joint} & \textbf{Both} \\ \hline
R &--&8.82&--\\ 
P & 9.58 & 8.49 & 8.49\\ 
H1 & 8.70 &  8.13 & 7.95\\ 
H2 & 8.12 &  8.04 & 7.66  \\ 
O1 & 8.19 & 7.92 & 7.86\\ 
O2 & 8.05 & 8.02 & 7.58\\ 
O3 & 7.87 & 7.75 & \textbf{7.48} \\ \hline
\end{tabular}
}
\end{table}

\begin{table*}[t]
\centering
\caption{Results of networks trained with only dense supervision (Dense) and both joint and dense supervision (Both) under various experimental settings on NYU dataset.}
\resizebox{0.95\textwidth}{!}{
\begin{tabular}{c|c|c|c|c|c|c}
\hline
\multirow{2}{*}{\textbf{Method}} & \multirow{2}{*}{\textbf{Backbone}} & \multirow{2}{*}{\textbf{Input Size}} & \multirow{2}{*}{\textbf{Dense Size}} & \multirow{2}{*}{\textbf{Params (FLOPs)}} & \multicolumn{2}{|c}{\textbf{Mean Error (mm)}} \\ \cline{6-7}
& & & & & \textbf{Dense} & \textbf{Both} \\ \hline
a & ResNet-18 & $128\times128$ & $64\times64$ & 15M (7.6G) & 8.53 & 7.75 \\
b & ResNet-50 & $128\times128$ & $64\times64$ & 34M (14.4G) & 7.87 & 7.48 \\
c & ResNet-101 & $128\times128$ & $64\times64$ & 53M (24.1G) & 7.81 & \textbf{7.37} \\ \hline
d & ResNet-50 & $128\times128$ & $32\times32$ & 33M (12.1G) & 8.07 & 7.68 \\
e & ResNet-50 & $128\times128$ & $16\times16$ & 32M (11.6G) & 8.24 & 7.85 \\
f & ResNet-50 & $96\times96$ & $48\times48$ & 34M (8.1G) & 8.14 & 7.59 \\ 
g & ResNet-50 & $96\times96$ & $24\times24$ & 33M (6.8G) & 8.32 & 7.76 \\ 
h & ResNet-50 & $96\times96$ & $12\times12$ & 32M (6.5G) & 8.37 & 8.01 \\ \hline
i & Hourglass-1stage & $128\times128$ & $64\times64$ & 4.6M(11.6G) & 8.38 & 7.70 \\ 
j & Hourglass-2stage & $128\times128$ & $64\times64$ & 8.7M(18.2G) & 7.73 & 7.43 \\ \hline

\end{tabular}
}
\end{table*}

\begin{small}
\begin{table}[t]
\centering
\caption{Results of three types of input modalities trained with only dense representation supervision (Dense) and both joint and dense representation supervision (Both).}
\resizebox{.8\columnwidth}{!}{
\begin{tabular}{c|c|c}
\hline
\multirow{2}{*}{\textbf{Input Modality}} & \multicolumn{2}{|c}{\textbf{Mean Error (mm)}} \\ \cline{2-3}
& \textbf{Dense} & \textbf{Both} \\ \hline
Voxels & 11.40 & 8.60 \\
Point clouds & 10.6 & 10.52 \\
Depth image & 7.87 & \textbf{7.48} \\ \hline
\end{tabular}
}
\end{table}
\end{small}

\begin{table}[t]
\centering
\caption{Impact on hyperparameters of offsets representation trained with only dense representation supervision (Dense) and both dense representation and joint supervisions (Both).}
\resizebox{1\columnwidth}{!}{
\begin{tabular}{c|c|c|c|c}
\hline
\textbf{Kernel Size} & 0.5 & 1 & 1.5 & 2\\ \hline
\textbf{Dense} & 8.3mm & 7.87mm & 7.95mm & 8.03mm \\ 
\textbf{Both} & 7.49mm & 7.48mm & 7.46mm & 7.45mm \\ \hline
\end{tabular}
}
\end{table}

\begin{table}[t]
\centering
\caption{Comparison with state-of-the-art methods on HANDS 2017 dataset. "SEEN" and "UNSEEN" denotes whether or not that the hand subject has been seen in the training set. "AVG" denotes all-joint mean error in millimeters over all test frames.}
\smallskip
\resizebox{.9\columnwidth}{!}{
\begin{tabular}{c|c|c|c}
\hline
\textbf{Method} & \textbf{SEEN} & \textbf{UNSEEN} & \textbf{AVG} \\ \hline
Vanora & 9.55 & 13.89 & 11.91 \\
THU VCLab & 9.15 & 13.83 & 11.70 \\
oasis & 8.86 & 13.33 & 11.30 \\
RCN-3D & 7.55 & 12.00 & 9.97 \\
V2V-PoseNet & 6.97 & 12.43 & 9.95 \\ 
A2J & 6.92 & 9.95 & 8.57 \\ \hline
Ours-ResNet18 & 5.76 & 9.57 & 7.84 \\
Ours-ResNet50 & \textbf{5.21} & \textbf{9.36} & \textbf{7.48} \\ \hline
\end{tabular}
}
\end{table}

\parskip=0pt
\subsection{Exploration Studies}

\textbf{Dense representation types.} As can be seen in Table 1, no matter what the dense representation is, aggregating operation based methods consistently outperform detection-based method, indicating that the operation can improve estimation accuracy through ensemble learning and it works effectively with all the representation types presented. Fig 3 shows that the proportion of good frames of AWR based methods outperforms detection-based methods at all thresholds, indicating that AWR significantly improves the network's robustness. Besides, we have several other observations. Firstly, aggregating operation based methods with only joint supervision outperforms directly regressing joint coordinates ($R$) thanks to the intermediate representation and adaptive weight maps. Secondly, since depth offsets can encode 3D geometric information of a 2.5D hand surface, depth offsets based methods consistently perform better than methods that directly predict absolute depth values ($H2 > H1$, $O2 > O1$). Finally, among these representations, 3D offsets ($O3$) performs the best since they encode rich spatial structure information and adopt more reasonable 3D distance measurement.

As shown in Table 1, directly performing dense joint regression is worse than directly regressing joint coordinates ($P\_Dense < R$). However, with the help of adaptive weighting, the performance is greatly improved ($P\_Joint > R$, $P\_Both > R$). We can come to two conclusions: 1) points at different positions possess different estimation accuracy, simply averaging the estimation of all points impairs the accuracy of final estimation ($P\_Dense < R$); 2) adaptively aggregating the results of different regions efficiently improves the estimation accuracy ($P\_Joint > R$), as both methods share almost the same network architecture except that $P\_Joint$ generates extra intermediate dense representation and use AWR to ensemble it.

\begin{figure}[t]
\centering
\includegraphics[width=1\columnwidth]{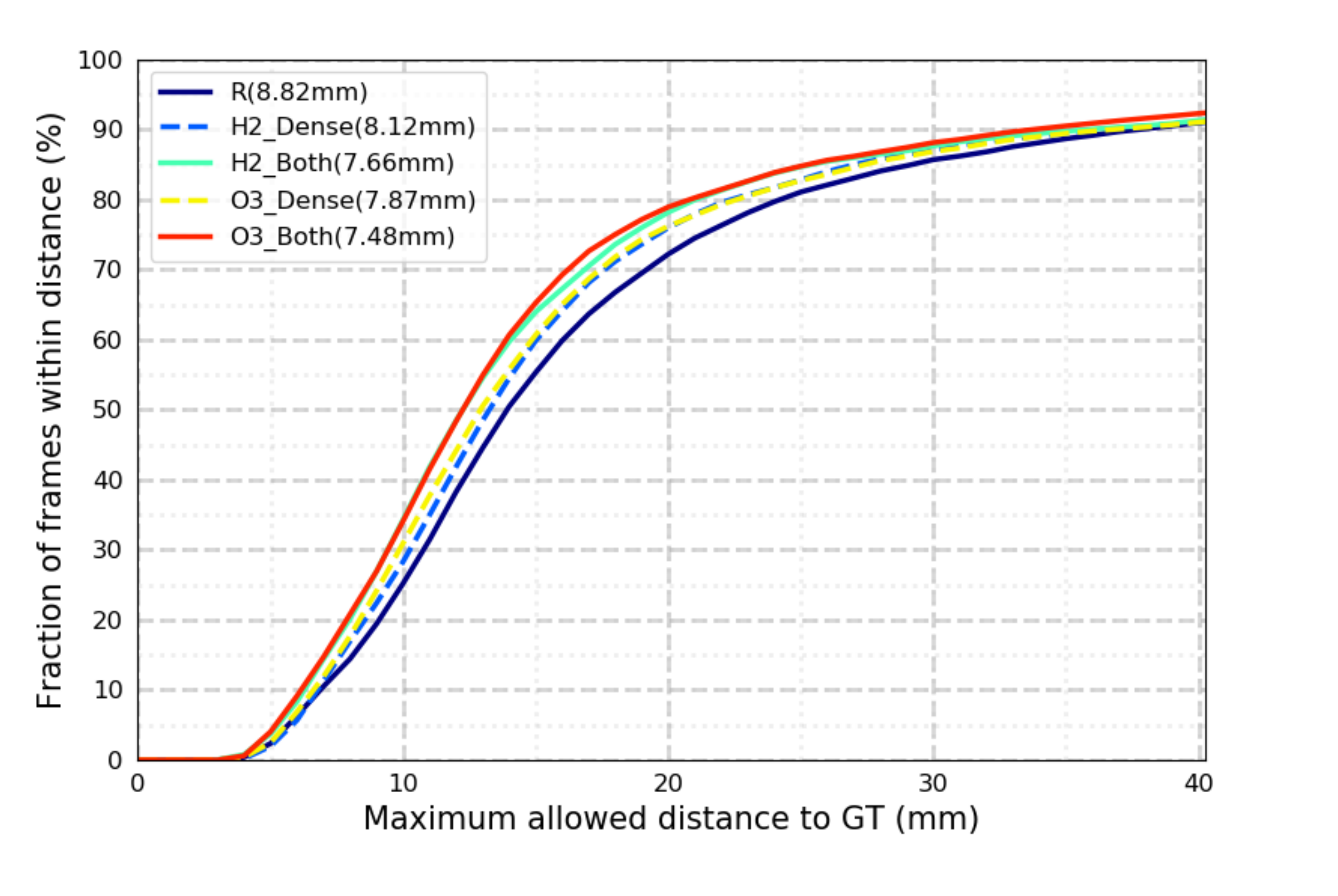} 
\caption{The proportion of good frames of directly regressing method (R), probability heatmaps with depth offset maps based method (H2) and 3D offsets (O3) based method using only dense representation supervision (Dense) or both dense representation and joint supervision (Both). Figure is best viewed in color.}
\end{figure}

\textbf{Input and dense representation sizes.} Results in Table 2 indicate that larger input and dense size obtain better accuracy. Method (f,d) and method (g,e) show that when input and dense size increase and decrease the same ratio respectively, the performance of AWR based method ($Both$) dropped while detection-based method ($Dense$) improved, reflecting that AWR based method is more sensitive to dense size than input size. We argue that since the accuracy of AWR based methods is related to the number of candidate points during aggregation, as dense size decreases, the number of candidate points drops drastically and impairs the estimation results. But generally, AWR based methods outperform detection-based methods with smaller input and output sizes, making our method favorable in real-world applications.

\textbf{Network architectures.} In this experiment, we first try out different network designs (ResNet \cite{resnet} and Hourglass \cite{hourglass}). As shown in Table 2, AWR outperforms method with only dense representation supervision and works well with single-stage 2D CNNs as well as cascaded network architecture (Method b, i, j), indicating that it can be easily embedded into different network architectures to boost the performance. For cascaded network, as stage number increases, performance gets better but at the cost of computational complexity. 

Then the depth of ResNet is experimented. Method a, b, c in Table 2 show that with the increase of network depths, mean joint error drops steadily for the two lines of work. Specifically, $a\_Both$ already achieves better results compared with $b\_Dense$ and $c\_Dense$ with far less of the computational and storage cost, indicating that simple network can obtain powerful performance through ensemble learning which further shows the superiority of our method. 

\textbf{Input modalities.} Three commonly used input modalities are experimented, including depth images, voxels and point clouds. As shown in Table 3, consistent improvements are brought by AWR under all three input modalities, demonstrating the effectiveness and generality of the aggregating operation under different coordinates and spaces.

We also experiment on the kernel size of 3D offsets, which is $k$ in Equation 3. As shown in Table 4, after applying AWR, the network is more robust to kernel size change than detection-based methods since introducing joint supervision brings weight maps certain degrees of adaptability. As kernel size increases, the accuracy of AWR method improved slightly. However, since larger kernel size requires more epochs and time to converge, to balance between training time and accuracy, we set kernel size as 1 in most experiments.

Table 1, 2 and 3 show that no matter what \textit{the representation types, network architectures, input and dense representation sizes or input modalities} are, AWR based methods consistently outperforms those trained with only dense representation supervision. Besides, AWR introduces robustness to hyperparameter of offsets representation, facilitating hyperparameter searching process. The results fully illustrate the effectiveness and generality of AWR and demonstrate the importance of ensemble learning.

\begin{figure*}[t]
\centering
\includegraphics[width=1\textwidth]{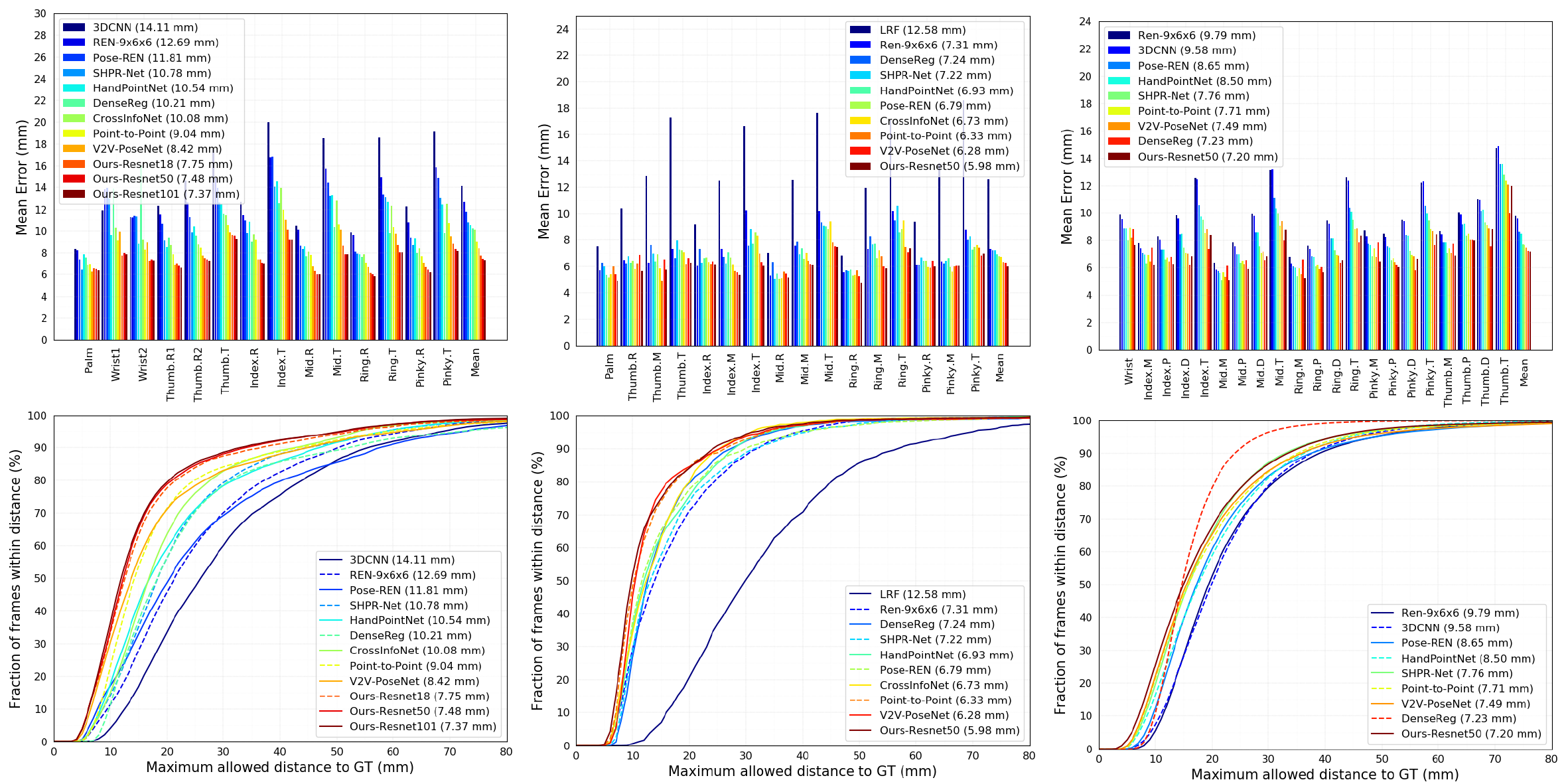} 
\caption{Comparison with state-of-the-art methods on NYU, ICVL and MSRA dataset. The all-joint and per-joint mean error (top row) and the proportions of good frames over different thresholds (bottom row)). Left: NYU dataset, middle: ICVL dataset, right: MSRA dataset. Figure is best viewed in color.}
\end{figure*}

\subsection{Comparison with State-of-the-art Methods}

We first compare our method with others on HANDS 2017 dataset \cite{hands17}. Since the HANDS 2017 dataset does not provide test set labels publicly, we evaluate using only mean joint error metric and compare our method with Vanora \cite{vanora}, THU VCLab \cite{pose}, oasis \cite{handpointnet}, RCN-3D \cite{vanora}, V2V-PoseNet \cite{v2v} and A2J \cite{a2j}. Results in Table 5 reflect that our ResNet18 based method already exceeds previous state-of-the-art methods by a large margin. And our ResNet50 based method further improves the average mean joint error by 0.36mm. The best result of mean joint error is 7.48mm. For seen and unseen hand objects, mean joint error are 5.21mm and 9.36mm respectively, reflecting the effectiveness and good generalization ability of our proposed method. Note that during comparison, we refer "Ours-ResNet18" and "Ours-ResNet50" as our method using ResNet18 and ResNet50 as backbone respectively.  

We also compare our method on NYU \cite{nyu}, ICVL \cite{icvl} and MSRA \cite{offset2} dataset with most of the state-of-the-art methods, including latent regression forest (LRF) \cite{lrf}, region ensemble network (REN-$9\times6\times6$) \cite{ren}, pose guided region ensemble network (Pose-REN) \cite{pose}, dense 3d regression (DenseReg) \cite{dense3d}, regresson-based 3d convolutional network (3DCNN) \cite{3dcnn}, voxel-to-voxel prediction network (V2V) \cite{v2v}, dense point-to-point regression pointnet (P2P) \cite{p2p}, pose estimation using point sets (HandPointNet) \cite{handpointnet}, point clouds based semantic hand pose regression network (SHPR-Net) \cite{shpr} and multi-task information sharing network (CrossInfoNet)\cite{crossinfonet}. 

As shown in Fig 4, our method outperforms all existing methods on the three 3D hand pose estimation datasets using either the per-joint and all-joint mean error or the proportion of good frames. Specifically, on \textbf{NYU dataset} \cite{nyu}, our ResNet18 based method already outperforms the superior V2V-PoseNet by 0.67mm with less storage comsumption and computational cost. For the proportion of good frames, all three of our methods outperforms other methods under all the thresholds. On \textbf{ICVL dataset} \cite{icvl}, our ResNet50 based method achieves better accuracy than other methods. When it comes to the proportion of good frames, it performs best when the threshold is smaller than 12mm and bigger than 20mm, and only slightly worse than V2V-PoseNet in between. Since our method uses cubes with a side length of 250mm to extract hand regions out of depth images, while V2V-PoseNet \cite{v2v} uses a smaller size of 200mm to locate hands more accurately. Therefore, a slight difference in performance is tolerable. On \textbf{MSRA dataset} \cite{offset2}, our method achieves comparable accuracy to DenseReg \cite{dense3d} and outperforms other methods. The proportion of good frames of our method is the best under 15mm threshold, but is relatively worse afterwards. However, as stated in \cite{fine}, part of the ground truth labels in ICVL \cite{icvl} and MSRA \cite{offset2} dataset are mistakenly annotated, which limits the learning capability of our method and leads to little improvement compared to other methods.

The comparison results show the superiority of our method in both accuracy and robustness over other state-of-the-art methods and proves that a simple baseline network can achieve powerful performance after introducing AWR.

\section{Conclusions}
We propose an adaptive weighting regression method for fast and robust 3D hand pose estimation from a single depth image. To overcome the existing defects of detection-based and regression-based methods and leverage the advantages of both, we aggregate different parts of the dense representation through discrete integration of all pixels in them to attain joint coordinates, guided by adaptive weight maps. This operation is learnable and can be easily embedded into different network architectures. It significantly improves the network's estimation accuracy and its robustness to situations where depth values around the target joint are missing and when there is severe self-similarity among fingers. We conduct comprehensive exploration experiments to illustrate the consistent improvement brought by AWR as well as its generality under various experimental settings. Powerful performance is obtained using simple baseline methods, making our method favorable in real-world applications. Our method achieves state-of-the-art performance on NYU, ICVL, MSRA and HANDS 2017 hand pose datasets. \cite{awr}

\bibliographystyle{aaai}
\bibliography{ref}
\end{document}